\documentclass{article}
\usepackage{times}
\usepackage{epsfig}
\usepackage{graphicx}
\usepackage{amsmath}
\usepackage{amssymb}
\usepackage{helvet} 
\usepackage{courier}  
\usepackage{graphicx}  
\usepackage{bm}
\usepackage{subfigure}
\usepackage{booktabs}
\usepackage{multirow}
\usepackage{caption}
\usepackage{float}
\usepackage{bbding}
\usepackage{array}
\usepackage{color}
\usepackage[pagebackref=true,breaklinks=true,letterpaper=true,colorlinks,breaklinks=true,bookmarks=false]{hyperref}

\usepackage[linesnumbered,ruled]{algorithm2e}
\usepackage{wrapfig}

    \usepackage[nonatbib,preprint]{neurips_2020}

\usepackage[utf8]{inputenc} 
\usepackage[T1]{fontenc}    
\usepackage{hyperref}       
\usepackage{url}            
\usepackage{booktabs}       
\usepackage{amsfonts}       
\usepackage{nicefrac}       
\usepackage{microtype}      

\title{ATSO: Asynchronous Teacher-Student Optimization for Semi-Supervised Medical Image Segmentation}

\usepackage{authblk}

\author{\textbf{Xinyue Huo}\textsuperscript{1}\thanks{This work was done when Xinyue Huo and Zijie Yang were interns at Huawei Noah's Ark Lab.}, \textbf{Lingxi Xie}\textsuperscript{2}, \textbf{Jianzhong He}\textsuperscript{2}, \textbf{Zijie Yang}\textsuperscript{3}, \textbf{Qi Tian}\textsuperscript{2}\\
\textsuperscript{1}University of Science and Technology of China,\quad\textsuperscript{2}Huawei Inc.,\quad\textsuperscript{3}Chinese Academy of Sciences\\
\texttt{xinyueh@mail.ustc.edu.cn},\quad\texttt{198808xc@gmail.com},\quad\texttt{jianzhonghe@pku.edu.cn},\\
\quad\texttt{yangzijie@ict.ac.cn},\quad\texttt{tian.qi1@huawei.com}
}

\date{} 

\begin{document}

\maketitle

\begin{abstract}
In medical image analysis, semi-supervised learning is an effective method to extract knowledge from a small amount of labeled data and a large amount of unlabeled data. This paper focuses on a popular pipeline known as self-learning, and points out a weakness named \textbf{lazy learning} that refers to the difficulty for a model to learn from the pseudo labels generated by itself. To alleviate this issue, we propose \textbf{ATSO}, an \textbf{asynchronous} version of teacher-student optimization. ATSO partitions the unlabeled data into two subsets and alternately uses one subset to fine-tune the model and updates the label on the other subset. We evaluate ATSO on two popular medical image segmentation datasets and show its superior performance in various semi-supervised settings. With slight modification, ATSO transfers well to natural image segmentation for autonomous driving data.
\end{abstract}

\section{Introduction}
\label{introduction}

Semantic segmentation plays an important role in medical image analysis. Recently, the fast development of deep learning~\cite{lecun2015deep} provides a powerful tool for dense image prediction~\cite{chen2018deeplab,long2015fully}, but for many scenarios of medical image analysis, data annotation is often expensive and thus difficult to acquire. This falls into the area of \textit{semi-supervised learning} which focuses on learning from both labeled data and unlabeled data, but the labeled part is often much smaller. An effective pipeline is known as \textit{self-learning}, in which an initial model is trained on the labeled part (training set) and fine-tuned on the unlabeled part (reference set) with the pseudo labels generated by itself.

We refer to this pipeline as \textit{teacher-student optimization}, a variant of knowledge distillation~\cite{hinton2015distilling} that has straightforward applications on medical image analysis~\cite{zhou2019semi}. However, we notice a factor that harms the efficiency of utilizing the unlabeled data. This is because in the continual learning procedure, the teacher and student models tend to have increasing similarity to each other. Consequently, the new supervision that the student model obtains from the pseudo labels becomes weaker and weaker, and the learning process quickly arrives at a plateau that is below satisfaction. This phenomenon is called \textbf{lazy learning} which, quantitatively, often reflects in that the quality of pseudo labels does not get improved during the learning process -- in other words, the accuracy on the reference set stops growing but the model itself does not know.

As the key observation of this paper, we point out that the essence of lazy learning is that the self-learning process is gradually pushing the teacher and student models, as a whole, towards a local optimum. In particular, the teacher model stores knowledge in the pseudo labels for the student model to learn; once a system error appears, it is likely to persist throughout the iteration. Therefore, inaccuracy accumulates and finally results in the unsatisfying quality of pseudo labels. To alleviate this problem, we present the \textbf{asynchronous teacher-student optimization} (ATSO) algorithm that proposes two modifications, both of which are designed to break up the chain of `error propagation'. \textbf{First}, we switch off continual learning and start each generation from the same initialized model. \textbf{Second}, we prevent using the pseudo labels generated by a teacher model to supervise its direct student. The second modification involves partitioning the reference set into two subsets. In each round of teacher-student optimization, we generate the pseudo labels on any subset based on a teacher model that was not trained on the same set of data. As we shall see in experiments, this strategy can largely improve the accuracy on the reference set and, consequently, on the unseen test set.

We perform experiments on the NIH and MSD datasets for pancreas segmentation from CT scans. Besides investigating semi-supervised learning within NIH dataset ($10\%$ or $20\%$ training data are labeled while others are not), we also investigate the issue of transfer learning (one dataset is fully annotated but the other is unlabeled, yet the goal is to achieve high accuracy on the second test set). In both cases, ATSO consistently outperforms the synchronous counterpart, as well as other semi-supervised learning methods by a significant margin. We also evaluate ATSO on Cityscapes and Mapillary, two autonomous driving datasets, showing its superiority in the domain of natural images.

\section{Asynchronous Teacher-Student Optimization}
\label{approach}

\subsection{Problem Setting, Fully-Supervised and Semi-Supervised Baselines}
\label{approach:baseline}

We study the problem of medical image segmentation in abdominal CT scans, and make use of 2D-based methods to process volumetric data. Each volume is partitioned into a set of slices along three directions, which are named the \textit{coronal}, \textit{sagittal}, and \textit{axial} views in the context of medical images. Each slice is then sent into a 2D network for segmentation, and the output is stacked into a 3D volume as the final prediction. From this point of view, the key problem is a standard semantic segmentation task and the evaluation protocol is called the Dice-S{\o}rensen coefficient (DSC), with the formula being ${\mathrm{DSC}}={2\times\left|\mathcal{Y}\cap\mathcal{Z}\right|/\left(\left|\mathcal{Y}\right|+\left|\mathcal{Z}\right|\right)}$, where $\mathcal{Y}$ and $\mathcal{Z}$ are the set of target voxels in the prediction and ground-truth, respectively. We choose the 2D segmentation baseline to be RSTN~\cite{yu2018recurrent}, an open-sourced, coarse-to-fine framework that achieved good performance in a few medical image segmentation tasks. Let us denote the network as ${\mathbf{y}}={\mathbf{f}\!\left(\mathbf{x};\boldsymbol{\theta}\right)}$, in which $\mathbf{x}$ and $\mathbf{y}$ denote the 2D input and output of the deep network $\mathbf{f}\!\left(\cdot\right)$ parameterized by weights $\boldsymbol{\theta}$.

Although RSTN as well as other algorithms~\cite{Isensee2019Automated,Xia2018Bridging,Perslev2019One} achieved satisfying performance under full supervision, the performance can degenerate dramatically in the scenarios of limited training data. For example, a fully-supervised RSTN ($60$ training samples) achieves an average accuracy of over $84\%$ on the NIH dataset, however, if only $10\%$ of training data ($6$ training samples) are preserved, the accuracy quickly drops to nearly $70\%$. This raises an important setting named semi-supervised learning, in which a large portion of training data do not have labels but we need to learn as much knowledge as possible from them. As a formal definition, the training set $\mathcal{T}$ is partitioned into two parts, namely, the supervised (labeled) set $\mathcal{S}$ and the reference (unlabeled) set $\mathcal{R}$. Most often, we have ${\left|\mathcal{S}\right|}\ll{\left|\mathcal{R}\right|}$. Also, there is a testing set, $\mathcal{E}$, which is invisible to the algorithm.

A simple and effective pipeline for semi-supervised learning is named self-learning. It starts with an initial model, denoted by $\mathbb{M}_0$, which is trained under supervised learning on $\mathcal{S}$. $\mathbb{M}_0$ gets updated for a total of $T$ rounds. In the $t$-th (${t}={1,2,\ldots,T}$ round (\textit{a.k.a.}, generation)), the reference subset, $\mathcal{R}$, is sent into the old model $\mathbb{M}_{t-1}$ (often referred to as the teacher model), and the prediction is named the pseudo label in the current round. The training process of the student model, $\mathbb{M}_t$, then follows a regular supervised learning procedure on both $\mathcal{T}$ and $\mathcal{R}$, with the supervision on $\mathcal{R}$ coming from the pseudo labels. The trained student model of the current round is used as the teacher model of the next round and the iteration continues till the end.

\subsection{Lazy Learning: the Devil in Self-Learning}
\label{approach:lazylearning}

We start with training the model using $10\%$ of data labeled and the remaining $90\%$ unlabeled on the NIH dataset. The detailed settings are elaborated in Section~\ref{semi:settings}. After $2$ generations, the self-learning pipeline achieves an accuracy of $78.98\%$ on the test set, claiming a significant improvement over $72.62\%$ that is reported by the model trained only on the $10\%$ labeled data. However, this number is still far below $85.08\%$ when the full labeled training set is available. That is to say, self-learning indeed extracts knowledge from the reference set, but the efficiency is far below satisfaction.

To understand the reason for this, we investigate the accuracy of the reference set, which is expected to grow with the learning procedure. However, we find that the accuracy quickly arrives at a plateau. After the $0$th (the initial training stage with only labeled data), $1$st, $2$nd, and $3$rd generations, the reference set accuracy is $71.41
\%$, $74.54\%$, $75.44\%$, and $74.72\%$, respectively. Compared to the scenario that $100\%$ training data have been labeled, the accuracy on the same subset of training data is $86.70\%$. In other words, the training procedure has entered a `trap' that the reference set stops at a low accuracy (what is worse, the accuracy starts to drop in the $3$rd generation), but the algorithm does not know because the ground-truth is missing.

From the viewpoint of optimization, this phenomenon can be explained as a local optimum of the self-learning system. Let $\mathbf{x}_n$ be a training sample in the reference set, $\mathcal{R}$, $\mathbf{y}_n^\star$ is the ground-truth label, and $\mathbf{y}_n$ is the predicted output by the teacher model. We assume that ${\mathbf{y}_n}={\mathbf{y}_n^\star+\boldsymbol{\epsilon}_n}$ where $\boldsymbol{\epsilon}_n$ is the prediction error. When $\mathcal{R}$ is labeled, \textit{i.e.}, $\mathbf{y}_n^\star$ is known, $\boldsymbol{\epsilon}_n$ follows a zero-mean distribution because the optimization goal is to minimize $\left|\boldsymbol{\epsilon}_n\right|$ on the training set. However, in the scenario of self-learning, $\mathbf{y}_n^\star$ remains unknown and thus $\boldsymbol{\epsilon}_n$ may follow a non-zero-mean distribution. Since $\mathbf{y}_n$ is used as the pseudo label, such noise can persist across in the student model. What makes things worse, each generation of the teacher-student optimization can introduce new noise and the noise can accumulate on the reference set. We use the name of \textbf{lazy learning} to refer to the behavior that \textit{the student model is unable to identify and eliminate the noise of the teacher model}.

\begin{figure}[!b]
\centering
\includegraphics[width=14cm,height=5cm]{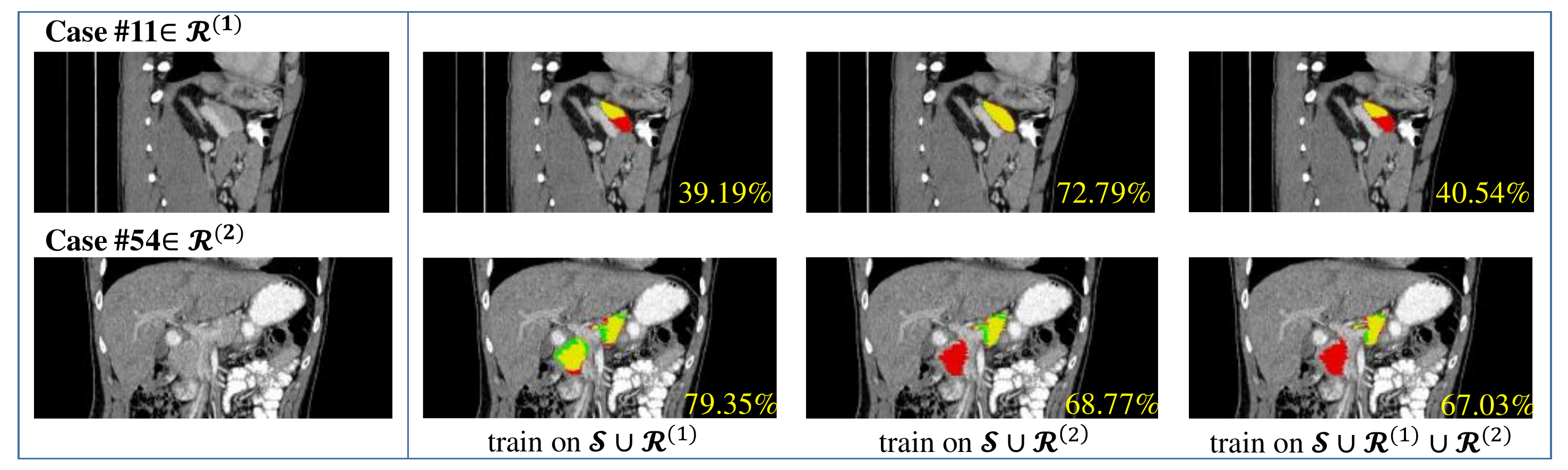}
\caption{Examples of lazy learning and a simple method to alleviate it \textit{(best viewed in color)}. The leftmost column shows two unlabeled images in the reference set, and the right columns show the segmentation results when the full reference set has been used for self-learning and when half of the reference set has been used. Segmentation accuracy is significantly improved when the reference set does not contain the test case. The \textcolor{red}{red}, \textcolor{green}{green}, and \textcolor{yellow}{yellow} masks indicate the true label, prediction, and overlapping region, respectively.}
\label{fig:lazy_learning}
\end{figure}

\subsection{Asynchronous Optimization: Escaping from the Trap}
\label{approach:solution}

To alleviate lazy learning, the key is to escape from the optimization trap, \textit{i.e.}, breaking up the chain that the noise, $\boldsymbol{\epsilon}_n$, is propagated from the teacher to the student. This is done from two aspects, both of which focus on \textbf{weakening the relationship between the teacher and student models}.

\textbf{First}, we prevent using the setting of continual learning that initializes the student model from the teacher model and performs fine-tuning, but starts training from the very beginning. To reveal the advantage, we notice that the optimization goal is to minimize $\left|\mathbf{y}_n^\star-\mathbf{f}\!\left(\mathbf{x}_n;\boldsymbol{\theta}\right)\right|$, but since $\mathbf{y}_n^\star$ is unknown, we use $\left|\mathbf{y}_n-\mathbf{f}\!\left(\mathbf{x}_n;\boldsymbol{\theta}\right)\right|$ to estimate it. The error of this estimation is bounded by $\left|\boldsymbol{\epsilon}_n\right|$:
\begin{equation}
{\left|\left|\mathbf{y}_n-\mathbf{f}\!\left(\mathbf{x}_n;\boldsymbol{\theta}\right)\right|-\left|\mathbf{y}_n^\star-\mathbf{f}\!\left(\mathbf{x}_n;\boldsymbol{\theta}\right)\right|\right|}\leqslant{\left|\mathbf{y}_n-\mathbf{y}_n^\star\right|}={\left|\boldsymbol{\epsilon}_n\right|}.
\end{equation}
That being said, provided that $\mathbf{y}_n^\star$ and $\mathbf{y}_n$ are fixed. Hence, when $\boldsymbol{\theta}$ is initialized independently from the teacher model, $\left|\mathbf{y}_n-\mathbf{f}\!\left(\mathbf{x}_n;\boldsymbol{\theta}\right)\right|$ is increased and thus the fraction that $\left|\boldsymbol{\epsilon}_n\right|$ occupies in the loss function is smaller. This effectively prevents the training procedure from being dominated by $\left|\boldsymbol{\epsilon}_n\right|$.

\textbf{Second}, we prevent using the same set of reference data continuously, \textit{i.e.}, always using the pseudo labels generated by a teacher model to supervise its direct student. To verify this assumption, we first notice that in the aforementioned self-learning procedure, the accuracy gained on the test set is much higher than that on the reference set, \textit{e.g.}, $6.36\%$ test accuracy gain vs. $4.03\%$ reference accuracy gain after $2$ self-learning generations.  To make things clearer, we partition the reference set into two parts, and perform self-learning on different combinations of reference data. Results of two hard examples are shown in Figure~\ref{fig:lazy_learning}. \textit{Interestingly, the segmentation accuracy is significantly improved when the example is not contained in the reference set.} This aligns with the observation that the test accuracy is higher than the reference accuracy. Back to the optimization perspective, \textit{the noise on the reference set does not transfer to the data that the current generation does not use for self-learning}. This inspires us to partition the reference set into two subsets, denoted by ${\mathcal{R}}={\mathcal{R}^{\left(1\right)}+\mathcal{R}^{\left(2\right)}}$. In each generation, we generate the pseudo labels on $\mathcal{R}^{\left(1\right)}$ using the model that was just self-trained on $\mathcal{R}^{\left(2\right)}$, and vice versa. After the last generation asynchronous update of both subsets, we combine the pseudo labels of both subsets into the complete reference set, based on which the final model is trained.


Integrating the above two aspects obtains the \textbf{asynchronous teacher-student optimization} (ATSO) algorithm that is described in Algorithm~\ref{alg:flowchart}. Of course, there exist other solutions that can alleviate lazy learning, yet our solution is simple and effective. Moreover, ATSO also has the option of dividing the reference set into more subsets, but this can slow down the training procedure. In practice, we find that using two folds of reference data performs well for semi-supervised learning.

\begin{algorithm}[!t]
\SetKwInOut{Input}{Input}
\SetKwInOut{Output}{Output}
\SetKwInOut{Return}{Return}
\Input{
a training set ${\mathcal{T}}={\mathcal{S}\cup\mathcal{R}}$ ($\mathcal{S}$ is labeled and $\mathcal{R}$ is unlabeled), \# of iterations $T$;
}
\Output{
a model $\mathbb{M}$ trained on $\mathcal{T}$;
}
Train an initial model $\mathbb{M}_0$ \textit{from scratch} on $\mathcal{S}$;\\
Divide $\mathcal{R}$ into two subsets, $\mathcal{R}^{\left(1\right)}$ and $\mathcal{R}^{\left(2\right)}$, ${t}\leftarrow{0}$, ${\mathbb{M}_0^{\left(1\right)}}\leftarrow{\mathbb{M}_0}$, ${\mathbb{M}_0^{\left(2\right)}}\leftarrow{\mathbb{M}_0}$;\\
\Repeat{${t}={T}$}{
Update $\mathcal{R}_{t+1}^{\left(1\right)}$ using the prediction from $\mathbb{M}_{t}^{\left(2\right)}$, $\mathcal{R}_{t+1}^{\left(2\right)}$ using the prediction from $\mathbb{M}_{t}^{\left(1\right)}$;\\
Train $\mathbb{M}_{t+1}^{\left(1\right)}$ \textit{from scratch} on $\mathcal{S}\cup\mathcal{R}_{t+1}^{\left(1\right)}$, train $\mathbb{M}_{t+1}^{\left(2\right)}$ \textit{from scratch} on $\mathcal{S}\cup\mathcal{R}_{t+1}^{\left(2\right)}$;\\
${t}\leftarrow{t+1}$;\\
}
Train $\mathbb{M}_{T}$ on $\mathcal{S}\cup\mathcal{R}_{t}^{\left(1\right)}\cup\mathcal{R}_{t}^{\left(2\right)}$;\\
\Return{
${\mathbb{M}}\leftarrow{\mathbb{M}_T}$.
}
\caption{ATSO: Asynchronous Teacher-Student Optimization}
\label{alg:flowchart}
\end{algorithm}

\section{Related Work}
\label{relatedwork}

\textbf{Medical image analysis} becomes increasingly important in assisting human doctors in clinics, in which semantic segmentation a fundamental basis of many tasks. This paper focuses on CT-scanned images, each of which is a 3D volume that corresponds to a specified region of the human body. Recently, with the fast development of deep neural networks~\cite{krizhevsky2012imagenet,simonyan2015very,he2016deep}, researchers developed effective algorithms for natural image segmentation~\cite{long2015fully,chen2018deeplab}, and these techniques quickly propagated to the area of medical images~\cite{ronneberger2015u,milletari2016v}. One of the major differences between segmenting natural and medical images lies in the dimensionality of input data. Medical scans often in 3D, but existing methods are often incapable of processing these data entirely. There are two popular pipelines, namely, 2D-based and 3D-based solutions, both of which sample data from the entire volume, process in the sampled region, and finally summarize all predictions into the original size. Differently, 2D-based pipelines~\cite{ronneberger2015u,roth2015deeporgan,zhou2017fixed,yu2018recurrent,man2019deep} sampled 2D slices, while 3D-based pipelines~\cite{cicek20163d,milletari2016v,zhu20183d} sample 3D sub-volumes. Both solutions have their advantages and disadvantages~\cite{tajbakhsh2016convolutional}, and researchers tried to integrate them into one framework to absorb benefits from both of them~\cite{liu20183d,ni2019elastic,Xia2018Bridging}.

\textbf{Semi-supervised learning} lies between supervised and unsupervised learning, which assumes that a small fraction of data are labeled, while the remaining part are unlabeled but closely related to the labeled subset~\cite{blum1998combining,zhu2005semi}. Researchers designed some generalized frameworks including self-learning~\cite{rosenberg2005semi}, multi-view learning~\cite{sun2013survey,xu2013survey}, co-training~\cite{blum1998combining}, \textit{etc}. The idea of \textbf{self-learning} is to use an initial model trained on labeled data to predict the labels on unlabeled data, so that these labels, though less accurate, can be used for training an updated, more powerful model~\cite{rosenberg2005semi}. This is related to knowledge distillation~\cite{hinton2015distilling} and teacher-student optimization~\cite{furlanello2018born}, but since unlabeled data was introduced, it is crucial to maximally improve the quality of the predicted label~\cite{tarvainen2017mean,yang2019training}. On the other hand, both \textbf{co-training} and \textbf{multi-view learning} aim to use the consistency within the task itself to assist learning. Differently, co-training often assumed that different models should produce the same output on the same data~\cite{qiao2018deep}, but multi-view learning assumed that the same model should produce the same result on various views of the same data~\cite{sun2013survey}. Sometimes, these assumptions were combined into one framework~\cite{sousa2017comparison,qiao2018deep}. Semi-supervised learning is of great interest to the researchers of medical image analysis, mainly because accurate annotations are often difficult to acquire. There exists large-scale datasets with inaccurate~\cite{wang2017chestx} and/or partial data annotations~\cite{zhou2019prior}, and researchers also developed practical semi-supervised algorithms for learning from these data~\cite{zhou2019semi,xia20203d}.

\textbf{ATSO} aims at improving the quality of `pseudo labels' in the self-training pipeline of semi-supervised learning. The key principle is to enlarge the difference between the teacher and student signals so that the student model learns non-trivial knowledge from the teacher model. A similar idea was presented by a recent work~\cite{yang2019snapshot} which studied fully-supervised learning tasks. Differently, \cite{yang2019snapshot} facilitates the difference by manipulating learning rates, while ATSO by isolating reference data between iterations.

\textbf{ATSO} is also related to other knowledge distillation approaches which trained a few models simultaneously so that each model can be used to supervise others. Examples include deep mutual learning~\cite{zhang2018deep} and deep co-training~\cite{qiao2018deep}. In particular, deep co-training~\cite{qiao2018deep} added an adversarial loss term to enlarge the gain between teacher and student models. Differently, we train two models individually on two subsets of the reference set, which naturally guarantees diversity and enjoys the ability of being parallelized when the number of individually-optimized models is large.

\section{Semi-Supervised Learning Experiments}
\label{semi}

\subsection{Datasets and Settings}
\label{semi:settings}

We evaluate ATSO on the NIH dataset~\cite{roth2015deeporgan} for pancreas segmentation. It contains $82$ normal CT scans, each of which is a 3D volume of $512\times512\times L$ voxels, where $L$ is the length of the long axis. We follow the prior work~\cite{zhou2017fixed,yu2018recurrent} to partition each dataset into four folds and use the first three folds ($62$ cases) as the training data. For semi-supervised learning, we follow the prior work~\cite{zhou2019semi,xia20203d} to use a small portion ($10\%$ or $20\%$) of labeled training data and leave the remaining part to the reference set. We report the average DSC over all test cases.

The configuration of the deep networks follows that in the original RSTN paper~\cite{yu2018recurrent}. In the training stage, we optimize three individual networks for segmentation along  with the \textit{coronal}, \textit{sagittal} and \textit{axial} views, respectively. In the inference stage, either on the reference set or the test set, predictions from these three views are fused into the final segmentation by majority voting.  Please refer to the 
original paper for further technical details.


\subsection{Quantitative and Qualitative Results}
\label{semi:within}

We first investigate semi-supervised segmentation on the NIH dataset with $10\%$ of the training set ($6$ cases) labeled. The naive baseline, by only using the labeled data for training, reports a $72.62\%$ test accuracy which is far lower than the upper-bound, $85.04\%$, when the labels of all training data are available. In what follows, we gradually add the key components of ATSO into the baseline and show how these components push the training procedure towards higher accuracy.

\noindent$\bullet$\quad\textbf{Breaking up Continual Learning Brings Significant Gain}

We first compare the options with and without continual learning, which we refer to as the \textbf{self-learning} baseline and \textbf{synchronous teacher-student optimization} (STSO), respectively. The former uses the last snapshot of the teacher model to initialize the student model, while the latter trains the student model from scratch. To save computational costs, we define the scratch model as the snapshot after the first two sections of RSTN, after which a randomly initialized transformation module is inserted that injects sufficient perturbations to break up the effect of continual learning.

Results are summarized in Table~\ref{tab:results_NIH}. The difference between the self-learning  baseline and STSO is significant. After two generations, the self-learning baseline achieves a $78.98\%$ accuracy on the testing set and $75.44\%$ on the reference set. Starting from the third generation, these numbers start to drop, demonstrating that lazy learning has obstructed the model from obtaining useful information. The best accuracy of STSO, $79.67\%$, is obtained after $4$ generations. That is to say, simply switching off continual learning leads to a non-trivial $0.69\%$ improvement and, as we shall see in Table~\ref{tab:comparison}, surpasses all the competitors.

\noindent$\bullet$\quad\textbf{Asynchronous Optimization Further Improves Accuracy}


\begin{wraptable}{r}{8cm}
\vspace{-0.3cm}
\centering
\resizebox{\linewidth}{16mm}{
\begin{tabular}{|c|c|c|c|c|c|c|}
\hline
\multirow{2}*{Generation} &
\multicolumn{2}{c|}{Self-learning} &\multicolumn{2}{c|}{STSO} &\multicolumn{2}{c|}{ATSO} \\

\cline{2-7}& @$\mathcal{R}$ 
& @$ \mathcal{E}$ & @$\mathcal{R}$ 
& @$ \mathcal{E}$ & @$\mathcal{R}$ 
& @$ \mathcal{E}$\\
\hline\hline {G0} & $71.41$ & $72.62$ &	$71.41$ & $72.62$  & $71.41$ & $72.62$  \\
\hline {G1} & $74.54$ & $76.82$ &	$74.54$ & 	$76.82$ & 	$75.69$  & $78.82$ \\
\hline {G2} & $75.44$ & $78.98$ &	$75.42$  & $77.88$ & 	$77.05$ & $80.81$ \\
\hline {G3} & $74.72$& $78.27$ &	$76.42$ & 	$79.27$ & 	$77.81$ & $\mathbf{81.69}$ \\
\hline {G4} & $74.38$ & $77.78$ &	$76.93$ & 	$79.67$ & 	$77.73$ & $81.41$ \\
\hline
{G5} & $73.38$ &  $77.22$ &	$77.15$ & 	$79.57$ & 	$\mathbf{78.07}$ & $81.57$ \\
\hline
\end{tabular} }
\caption{Segmentation results (DSC, $\%$) on the NIH pancreas segmentation datasets with $10\%$ labeled training data ($6$ cases). The results of the reference set and the test set are compared during $5$ generations.}
\label{tab:results_NIH}
\vspace{-0.3cm}
\end{wraptable}

Next, we study the difference between STSO and ATSO. Results are summarized in Table~\ref{tab:results_NIH}. ATSO improves segmentation accuracy on both the reference and test sets. Detailed results on the reference subsets during iteration are provided in Appendix~\ref{extra1}. Interestingly, ATSO enjoys faster growth in both numbers: after only two generations, the test accuracy has increased to over $80\%$, claiming a nearly $3\%$ advantage over the corresponding number of STSO. After five generations, ATSO still enjoys a $2\%$ advantage over STSO. That being said, ATSO has a broad range of applications in the scenario of limited computational resource for model training.

More importantly, these results indicate that the lazy learning phenomenon indeed persists during the entire training process. Therefore, \textit{by not generating pseudo labels on the reference set that was just used}, the algorithm can escape from the optimization trap. We notice that compared to STSO that always refers to the entire reference set, ATSO trains each model using only half of the reference data, which has a natural deficit but still achieves higher accuracy. We expect ATSO to have a larger advantage when the amount of unlabeled data becomes larger.

\noindent$\bullet$\quad\textbf{Comparison to the State-of-the-Arts, and Visualization}

In Table~\ref{tab:comparison}, we compare ATSO against state-of-the-art approaches, and show that ATSO outperforms all of them. In particular, ASTO surpasses~\cite{xia20203d} by more than $2.5\%$ in both scenarios that $10\%$ and $20\%$ labeled data have been used. Note that~\cite{xia20203d} is a recently published method which involved uncertainty in multi-view learning -- in comparison, our solution is easier and more effective. In addition, being simple and easily implemented, ATSO can be combined with other training strategies, \textit{e.g.}, adversarial training~\cite{qiao2018deep} or uncertainty evaluation~\cite{xia20203d}, towards better performance.

\begin{figure*}[!t]
\centering
\includegraphics[width=\textwidth]{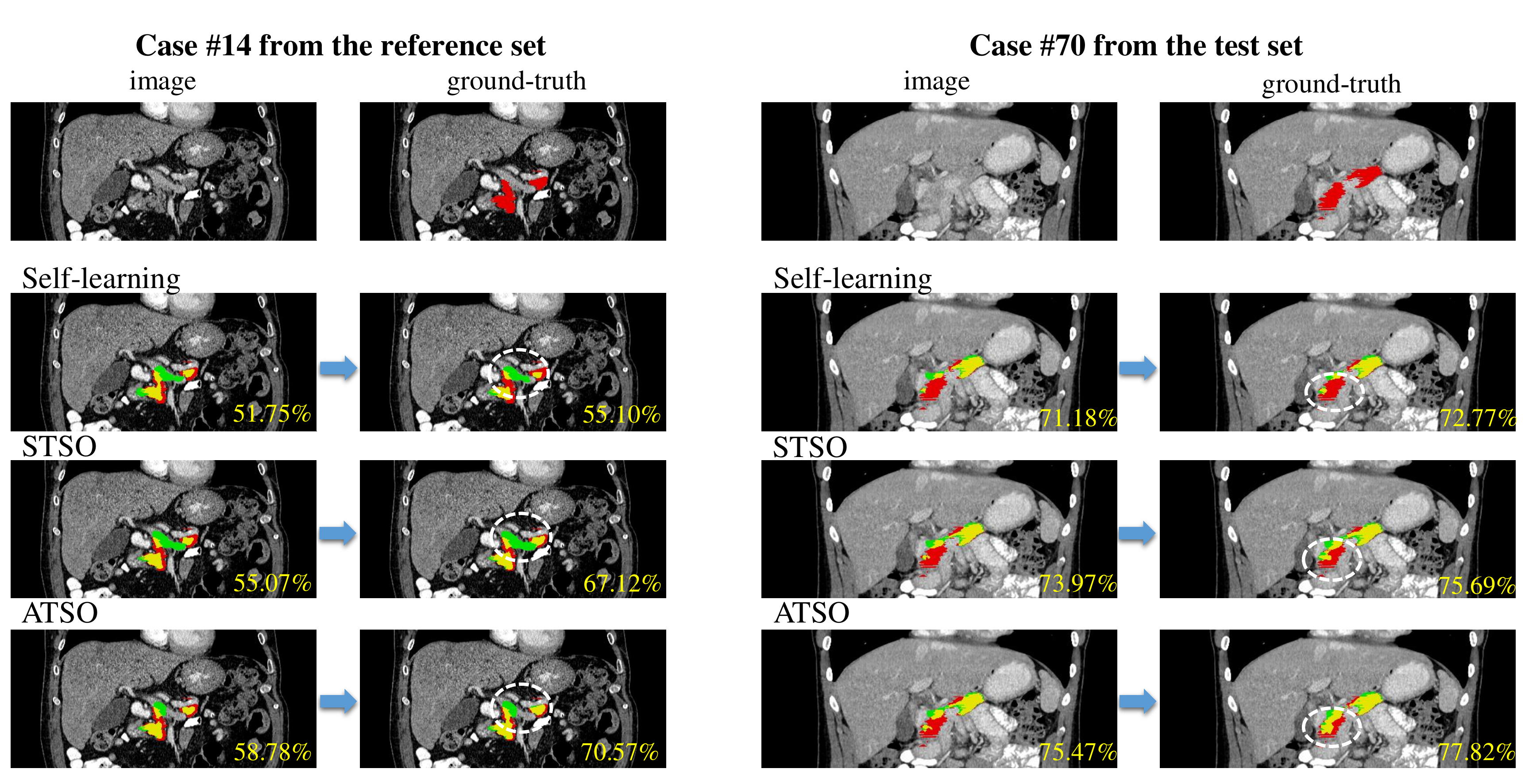}
\caption{Visualization of the improvement on the reference and test sets (best viewed in color). For each case, the second to the last row show the results produced by the self-learning baseline, STSO, and ATSO, respectively. In each pair, the left and right sides of the arrow are the outputs of an intermediate and the final generations. We show typical 2D slices that reflect the difference, while the DSC numbers in the bottom-right corner are computed in the entire 3D volume. The \textcolor{red}{red}, \textcolor{green}{green}, and \textcolor{yellow}{yellow} masks indicate the true label, prediction, and overlapping region, respectively. Please also zoom in to see the 
white dashed circles that mark the regions with significant accuracy gain.}
\label{fig:visualization}
\end{figure*}

Figure~\ref{fig:visualization} shows some typical examples of how segmentation errors are fixed with semi-supervised learning. When the labeled set is small, it is very likely that the labeled training set does cover sufficient situations, causing some failure cases in the reference set. In the self-learning baseline or even STSO, it is relatively difficult for the model to fix these errors during iteration, and the persisted errors can hinder the ability of the student model. In comparison, ATSO offers extra opportunities to jump out of the current distribution and thus get rid of the failure case. Hence, the efficiency of utilizing unlabeled training data is improved.

\begin{table}[!b]
\centering
\small
\setlength{\tabcolsep}{0.08cm}
\resizebox{\textwidth}{12mm}{
    \begin{tabular}{|l|l|r|r||l|l|r|r|}
    \hline
    Method & Backbone  &  $10\%$ D &  $20\%$ D & Method & Backbone  &  $10\%$ D &  $20\%$ D \\
    \hline\hline
    DMPCT~\cite{zhou2019semi}             & 2D ResNet-101       &          $63.45$ &$66.75$ & UMCT~\cite{xia20203d} ($2$v fusion)   & 3D ResNet-18        &          $77.78$ &          $80.52$ \\ 
    \hline
    DCT~\cite{qiao2018deep} ($2$v)        & 3D ResNet-18        &          $71.43$ &  $77.54$ & UMCT~\cite{xia20203d} ($3$v fusion)   & 3D ResNet-18        &          $79.05$ &          $81.18$ \\
    \hline TCSE~\cite{li2018semi}                & 3D ResNet-18   & $73.87$ &          $76.46$ & Self-Learning (ours)      & 2D FCN8s $\times2$  &          $78.98$ &          $82.87$ \\
    \hline UMCT~\cite{xia20203d} ($2$v)          & 3D ResNet-18        &          $75.63$ &          $79.77$  & STSO (ours)  & 2D FCN8s $\times2$  &          $79.67$ &          $83.21$ \\
    \hline UMCT~\cite{xia20203d} ($6$v)          & 3D ResNet-18        &          $77.87$ &          $80.35$  & ATSO (ours)       & 2D FCN8s $\times2$  & $\mathbf{81.69}$ & $\mathbf{83.70}$ \\
    \hline
    \end{tabular} }
\caption{Accuracy (DSC, $\%$) comparison between some recently published methods and our solutions, \textit{i.e.}, the self-learning baseline, STSO, and ATSO. We have tested the accuracy using either $10\%$ or $20\%$ labeled training data. Some of the numbers are borrowed from~\cite{xia20203d}. $2$v means that $2$ views have been used in multi-view learning.}
\label{tab:comparison}
\end{table}

\section{Transfer Learning Experiments}
\label{transfer}

Beyond semi-supervised learning, we consider another scenario that has a broad range of applications, \textit{i.e.}, transferring the segmentation model trained on one dataset to another that does not have labels at all. For this purpose, we start with medical image segmentation and later generalize to natural images where a larger number of semantic classes are present.

\subsection{Transfer Segmentation from NIH to MSD}
\label{transfer:medical}

The pancreatic tumor subset of the MSD dataset (\textsf{http://medicaldecathlon.com/}) has $281$ abnormal CT scans. Each sample contains the annotation of the pancreas and the tumor. Since we train the base model on the NIH dataset (normal pancreas), we only evaluate the pancreas segmentation accuracy on the MSD data. Transferring from NIH to MSD (the NIH training data are labeled while the MSD training data are not) is a challenging task since the distributions are very different between these two datasets: the scanners are different, and the MSD data contain abnormality (pancreatic tumor) while the NIH data do not. In this scenario, the key is to extract information from the unlabeled training set that has a close distribution to the test set. Since the resolution on the long axis varies significantly, we normalize the inter-slice distance along the long axis during training and testing, but rescale the final output to the original size for a fair comparison. A global DSC (combining all test cases into a single volume) is computed.

Directly applying the pre-trained model for segmentation reports an average accuracy of $69.95\%$ on the test set. After four generations, ATSO reports an average accuracy of $76.71\%$ (an over $6.76\%$ gain over the direct transfer baseline) which is higher than that of STSO ($75.48\%$) and the self-learning baseline ($74.78\%$) after the same number of generations. Note that $76.71\%$ is even higher than the accuracy ($75.49\%$) obtained from using $62$ labeled MSD data for training. Similar to the NIH experiments, we also obtain more accurate pseudo labels. After four generations, ATSO achieves a $71.67\%$ accuracy on the reference set, but both the self-learning baseline and STSO reports around $70\%$. This aligns with our expectation: ATSO has the ability of transferring knowledge from a labeled dataset to another unlabeled, even when the data distributions differ considerably. Detailed results of transfer segmentation experiments from NIH to MSD are provided in Appendix~\ref{extra2}.

\subsection{Transfer Segmentation across Natural Image Datasets}
\label{transfer:natural}

The second transfer learning scenario is defined between two natural image segmentation datasets. We use two popular datasets for autonomous driving, \textit{i.e.}, training the model on Cityscapes~\cite{7780719} and transferring it to Mapillary (\textsf{https://www.mapillary.com/})\footnote{As a side comment, we also investigate the setting of semi-supervised learning within the Cityscapes dataset, and provide the results in Appendix~\ref{extra3}. The conclusions are similar to that reported in the NIH experiments.}. The labeled training set contains $2\rm{,}975$ finely annotated images from Cityscapes and the reference and test sets are from Mapillary, which have $18\rm{,}000$ and $2\rm{,}000$ images, respectively. Note that Cityscapes has $19$ semantic classes while Mapillary has $66$. We train a model to infer $19$ classes and evaluate it on the reduced ground-truth on Mapillary by map each of the $66$ classes to one of the $19$ classes. The direct transfer method reports a mIOU of $24.70\%$ on the Mapillary test set that is dramatically lower than the number ($77.86\%$) on the Cityscapes test set, revealing a significant distribution gap between these two datasets.

We directly apply STSO and ATSO in this scenario, and obtain mIOU values of $28.11\%$ and $28.26\%$ (see Table~\ref{tab:comparison_transfer}), respectively. ATSO does not show advantages over STSO. This is mainly because $19$-class segmentation is more challenging, so the noise in the pseudo labels may have been magnified to overwhelm the difference between STSO and ATSO. As an example, for the objects with a small ground-truth area (\textit{e.g.}, a \textit{train}), the teacher model (not seeing the ground-truth) can easily ignore the entire object so that the pseudo label will guide the student models to make the same critical mistake. To alleviate this instability, we propose a solution that weakens the pseudo labels to get rid of the trouble. We further perform a pre-defined mapping to reduce the number of classes from $19$ to $5$ (\textit{i.e.}, \textit{rode}, \textit{vehicle}, \textit{person}, \textit{vegetation}, \textit{others}) so as to reduce the risk of missing some class completely. In this reduced, $5$-class segmentation task, ATSO reports a mIOU of $77.11\%$ which is significantly higher than the mIOU produced by STSO, $69.94\%$.

We then use the $5$-class pseudo label, generated by the final generation of either STSO or ATSO, to guide the target network. During the training procedure, the target network generates $19$-class segmentation and then reduces it to $5$ classes for computing the loss function with respect to the pseudo labels. With only one training generation, the target model produces mIOU of $29.76\%$ and $34.36\%$ using the pseudo labels from STSO and ATSO, respectively, \textit{i.e.}, better pseudo labels lead to higher mIOU. Interestingly, when we continue fine-tuning the latter model (a $34.36\%$ mIOU) \textbf{with $19$-class pseudo labels}, its performance is further boosted to $37.23\%$, claiming an over $12\%$ gain beyond the direct transfer baseline ($24.70\%$). Class-wise IOU numbers are detailed in Table~\ref{tab:comparison_transfer}. These results provide new insights to apply semi-supervised segmentation to the challenging datasets, in particular, with difficult semantic classes that can be easily missed. More details and results of the transfer experiments can be found in Appendix~\ref{extra4}.

\begin{table}[!t]
\centering
\small
\setlength{\tabcolsep}{0.08cm}
\resizebox{\textwidth}{10mm}{
\begin{tabular}{|l|l|l|l|l|l|l|l|l|l|l|l|l|l|l|l|l|l|l|l|l|l|}
\hline
Method & \textit{road} & \textit{side} & \textit{budg} & \textit{wall} & \textit{fence} & \textit{pole} & \textit{tr lt} & \textit{tr sn} & \textit{vegtr} & \textit{terr} & \textit{sky} & \textit{pers} & \textit{rider} & \textit{car} & \textit{truck} & \textit{bus} & \textit{train} & \textit{motor} &  \textit{bike} & \textbf{mIoU} \\
\hline
transfer & 61.82 &  10.23 & 46.98 &4.73 & 13.98 & 17.48 & 16.66 & 19.08 & 64.53 & 32.09 & 54.58 & 25.51 & 7.88 & 54.46 & 9.36 & 10.20 & 0.07 & 10.44 & 9.51 & 24.70  \\
\hline\hline
$\mathrm{STSO}_{19}$ & 60.11 &  7.68 & 57.64 &0.57 & 10.18 & 10.73 & 14.11 & 24.77 & 71.50 & $\mathbf{44.88}$ & 54.88 & 43.54 & 7.77 & 59.68 & 6.79 & 20.82 & 0.02 & 21.28 & 17.18 & 28.11  \\
\hline
$\mathrm{ATSO}_{19}$ & 62.08 &  9.64 & $\mathbf{62.46}$ &0.90 & 8.79& 6.84 & 15.90 & 26.02 & 71.50 & 35.48 & $\mathbf{63.36}$ & 45.61 & 15.89 & 66.44 & 5.83 & 15.89 & 0.05 & 14.47 & 27.41 & 28.26  \\
\hline\hline
$\mathrm{STSO}_{5}$& 69.87 &9.60 & 40.31 &14.42 & 14.78 & $\mathbf{26.36}$ & 20.90 & 21.49 & 83.28 & 25.94 & 27.15 &51.27 & 9.11 & 80.14 & 11.60 &11.15 & 0.43 & 22.16 &25.48 & $\mathbf{29.76}$  \\ 
\hline
$\mathrm{ATSO}_{5}$ & $\mathbf{80.65}$ &  $\mathbf{12.07}$ & 40.24 & $\mathbf{16.67}$ & 21.62 & 25.40 & $\mathbf{31.88}$ & $\mathbf{35.51}$ & $\mathbf{86.07}$ & 16.75& 30.09 & 56.83 & 14.27 & 82.91 & 23.25 & 15.78 & 2.22 & 28.09 & 32.68 & 34.36  \\
\hline
$\mathrm{ATSO}_{5\rightarrow19}$& 80.01 &  7.37 & 42.87 & 13.64 & $\mathbf{24.69}$ & 19.82 & 23.46 & 20.03 & 84.81 & 34.79 & 19.72 & $\mathbf{60.86}$ & $\mathbf{29.43}$ &$\mathbf{83.20}$ & $\mathbf{24.56}$ & $\mathbf{27.12}$ & $\mathbf{15.64}$ & $\mathbf{42.38}$ & $\mathbf{49.17}$ & $\mathbf{37.23}$  \\
\hline
\end{tabular} 
}
\caption{Class-wise and mean IOU (\%) of Mapillary, produced by different training strategies. Please refer to the texts for the detailed information of these models. Please zoom in to see clearer.}
\label{tab:comparison_transfer}
\end{table}

\section{Conclusions}
\label{conclusions}

In this paper, we investigate the problem of semi-supervised learning, in particular, using teacher-student optimization, in the scenario of medical image segmentation. The core discovery is that the self-learning process can fall into a trap named lazy learning which downgrades the quality of prediction in the reference set. To alleviate this issue, we propose a simple yet effective pipeline named \textbf{asynchronous teacher-student optimization} (ATSO) which (i) switches off continual learning and (ii) avoids any unlabeled sample to be used in two consecutive fine-tuning rounds. Experiments on a few public datasets verify the effectiveness of our approach in both intra-dataset and inter-dataset semi-supervised learning tasks.

Our research sheds light on a new direction to improve semi-supervised learning, \textit{i.e.}, design a better schedule of feeding unlabeled data to the model. We expect to generalize this methodology to a wider range of learning tasks, including those on natural images. Also, the ATSO method needs further investigation, in particular, to solve the problem that unlabeled training data still suffer lower accuracy than the unseen testing data, as observed in all experiments.

\section*{Broader Impact}

This paper presents a novel algorithm to improve the accuracy of semi-supervised medical image segmentation. We summarize the potential impact of our work in the following aspects.

\begin{itemize}
\item \textbf{To the research community.} We reveal an interesting phenomenon named lazy learning, and propose an asynchronous optimization method to alleviate it. This provides a new insight to the semi-supervised learning community. In particular, conventional approaches mainly focused on introducing priors to make use of the unlabeled data, and we point out that such prior can sometimes lead us to a local optimum. Our solution can be integrated with many self-learning algorithms.
\item \textbf{To training with limited labeled data.} In the current era, a large amount of data of different types and modalities pour in everyday. However, few of them have been annotated. It has become increasingly important for the AI algorithms to learn from unlabeled data, and our algorithm advances the research in this field. Intriguingly, the ATSO method can be applied to different kinds of image data, including medical scans and autonomous driving data.
\item \textbf{To the downstream engineers.} With two examples (medical image and street image segmentation) studied in the paper and our code released, it will become easier for engineers to deploy our algorithm to other scenarios. While this may help to develop AI-based applications, there exist risks that some engineers, with relatively less knowledge in deep learning, can deliberately use the algorithm, \textit{e.g.}, without considering the data distribution of the reference set, which may actually harm the performance of the designed system.
\item \textbf{To the society.} There is a long-lasting debate on the impact that AI can bring to the human society. In particular, applying AI algorithms to medical care has become increasingly popular recently. Our algorithm can assist constructing a better system for medical image analysis, which is important to the scenarios that human expertise is limited or expensive, or in a public health event like the pandemic of COVID-19. However, we also notice that our algorithm can help to use personal medical data even if most of them have not been annotated. This raises concerns on privacy. Therefore, in general, it can bring both beneficial and harmful impacts and it really depends on the motivation of the users.
\end{itemize}

We also encourage the community to investigate the following problems.
\begin{enumerate}
\item Is there any better solution for asynchronous optimization? The current method suffers the problem that being able to use only half of reference data in each generation.
\item Are there any deeper insights of lazy learning? Is it related to some intrinsic difficulties of optimization that were not studied in this paper?
\item Is it possible to generalize this algorithm to other vision tasks (\textit{e.g.}, image classification, object detection, \textit{etc.}) or a generalized machine learning scenario?
\end{enumerate}

{\small
\bibliographystyle{plain}
\bibliography{egbib}
}

\appendix

\section{Additional Results of Semi-Supervised Learning on NIH}
\label{extra1}

This part corresponds to Section~\ref{semi:within} of the main article. The detailed results regarding the segmentation accuracy on the two reference subsets are summarized in Table \ref{tab:NIH_subset}. In each generation, we evaluate the two models on both $\mathcal{R}^{\left(1\right)}$ and $\mathcal{R}^{\left(2\right)}$. We can see that in the first two generations, the model that gets trained on the other reference subset performs better, which verifies the \textit{lazy learning} phenomenon. As the training process continues, this phenomenon becomes less significant, and this may be the main reason for the shrunk advantage of ATSO over STSO after the first two stages. In addition, this motivates us to consider if there is a better way to make use of data for the ATSO algorithm.

\begin{table}[!h]
\centering
\setlength{\tabcolsep}{0.08cm}
\begin{tabular}{|c|c|c|c|c|c|c|}
\hline
Generation &
$\mathbb{M}_{t}^{\left(1\right)}$ @ $\mathcal{R}^{\left(1\right)}$ &
$\mathbb{M}_{t}^{\left(2\right)}$ @ $ \mathcal{R}^{\left(1\right)}$ &
$\mathbb{M}_{t}^{\left(1\right)}$ @ $ \mathcal{R}^{\left(2\right)}$ &
$\mathbb{M}_{t}^{\left(2\right)}$ @ $ \mathcal{R}^{\left(2\right)}$ & updated $     \mathcal{R}_t    $ & 
$\mathbb{M}_{t} $ @ $ \mathcal{E}$ \\
\hline\hline
G0 & \multicolumn{2}{c|}{$70.85$} 
 & \multicolumn{2}{c|}{$71.96$} 
 & $71.41$ & $72.62$\\
 \hline
G1 & $72.30$ &$75.61$
 & $75.76$ & $75.57$
 & $75.69$ & $78.82$\\
  \hline
G2 & $77.33$ &$76.42$
 & $77.67$ & $76.97$
 & $77.05$ & $80.81$\\
  \hline
G3 & $77.74$ &$77.32$
 & $78.30$ & $78.59$
 & $77.81$ & $81.69$\\
  \hline
G4 & $78.16$ &$76.77$
 & $78.69$ & $79.28$
 & $77.73$ & $81.41$\\
  \hline
G5 & $77.61$ &$77.47$
 & $78.67$ & $79.71$
 & $78.07$ & $81.57$\\
\hline
\end{tabular} 
\caption{Detailed results (segmentation accuracy, $\%$) on two reference subsets in ATSO. This experiment is done with $10\%$ of training data labeled on the NIH dataset.}
\label{tab:NIH_subset}
\end{table}

\begin{table}[!h]
\centering
\setlength{\tabcolsep}{0.08cm}
\begin{tabular}{|c|c|c|c|c|c|c|}
\hline
Generation &
$\mathbb{M}_{t}^{\left(1\right)}$ @ $\mathcal{R}^{\left(1\right)}$ &
$\mathbb{M}_{t}^{\left(2\right)}$ @ $ \mathcal{R}^{\left(1\right)}$ &
$\mathbb{M}_{t}^{\left(1\right)}$ @ $ \mathcal{R}^{\left(2\right)}$ &
$\mathbb{M}_{t}^{\left(2\right)}$ @ $ \mathcal{R}^{\left(2\right)}$ & updated $     \mathcal{R}_t    $ & 
$\mathbb{M}_{t} $ @ $ \mathcal{E}$ \\
\hline\hline
G0 & \multicolumn{2}{c|}{$60.30$} 
 & \multicolumn{2}{c|}{$67.19$} 
 & $63.75$ & $69.95$\\
 \hline
G1 & $64.12$ &$63.60$
 & $74.11$ & $72.11$
 & $68.86$ & $74.53$\\
  \hline
G2 & $62.92$ &$64.99$
 & $73.35$ & $74.17$
 & $69.17$ & $76.57$\\
  \hline
G3 & $65.50$ &$65.59$
 & $75.39$ & $74.75$
 & $70.49$ & $76.10$\\
  \hline
G4 & $64.75$ &$66.07$
 & $75.97$ & $75.73$
 & $71.02$ & $76.71$\\
\hline
\end{tabular} 
\caption{Detailed results (segmentation accuracy, $\%$) of two reference subsets in ATSO. This transfer experiment is done using all labeled training data on NIH and all unlabeled training data on MSD.}
\label{tab:MSD_subset}
\end{table}

\section{Additional Results of Transfer Learning from NIH to MSD}
\label{extra2}

This part corresponds to Section~\ref{transfer:medical} of the main article. The detailed results regarding the segmentation accuracy on the two reference subsets are summarized in Table \ref{tab:MSD_subset}. Conclusions are very similar to that observed in the semi-supervised experiments on NIH.

We also compare the segmentation results on the reference and test sets among the self-learning baseline, STSO and ATSO. Results are summarized in Table~\ref{tab:compare_transfer}. We can observe that the performance of the self-learning baseline begins to degrade after two generations, while that of STSO and ATSO does not.

\begin{table}[!h]
\centering
\begin{tabular}{|c|c|c|c|c|c|c|}
\hline
\multirow{2}*{Generation} &
\multicolumn{2}{c|}{Self-learning} &\multicolumn{2}{c|}{STSO} &\multicolumn{2}{c|}{ATSO} \\

\cline{2-7}& @$\mathcal{R}$ 
& @$ \mathcal{E}$ & @$\mathcal{R}$ 
& @$ \mathcal{E}$ & @$\mathcal{R}$ 
& @$ \mathcal{E}$\\
\hline\hline {G0}  & $63.75$ & $69.95$	& $63.75$ & $69.95$ & $63.75$ & $69.95$  \\
\hline {G1} & $68.24$ & $75.35$ &	$68.24$ & 	$75.35$ & 	$68.86$  & $74.53$ \\
\hline {G2} & $69.72$ & $75.42$ &	$68.50$  & $75.09$ & 	$69.17$ & $76.57$ \\
\hline {G3} & $69.77$& $74.78$ &	$68.04$ & 	$74.17$ & 	$70.49$ & $76.10$ \\
\hline {G4} & $70.21$ & $74.03$ &	$69.08$ & 	$75.48$ & 	$\mathbf{71.02}$ & $\mathbf{76.71}$ \\
\hline
\end{tabular} 
\caption{Segmentation results (DSC,$\%$) on the transfer learning  from NIH to MSD. The results of the reference set and the test set are compared during  $4$ generations.}
\label{tab:compare_transfer}
\end{table}

\section{Additional Results of Semi-Supervised Learning on Cityscapes}
\label{extra3}

This part corresponds to Section~\ref{transfer:natural} of the main article. We investigate the semi-supervised learning on the natural images for autonomous driving. We perform experiments on the Cityscapes dataset with $1\mathrm{K}$ finely-annotated images used for training and the remaining part of training data ($21\rm{,}944$ images, including finely-labeled and coarsely-labeled images) used for reference.

Results are summarized in Table~\ref{tab:comparison_cityscapes}. We still compare among four algorithms, namely, only training on the labeled data (transfer), the self-learning baseline, STSO, and ATSO. We can observe the same phenomenon as in the semi-supervised learning experiments in the NIH dataset, where ATSO improves consistently upon the other competitors. In particular, the overall improvement is over $15\%$, and the improvement on hard classes is more significant.

\begin{table}[!h]
\centering
\setlength{\tabcolsep}{0.08cm}
\resizebox{\textwidth}{8mm}{
\begin{tabular}{|l|l|l|l|l|l|l|l|l|l|l|l|l|l|l|l|l|l|l|l|l|l|}
\hline
Method & \textit{road} & \textit{side} & \textit{budg} & \textit{wall} & \textit{fence} & \textit{pole} & \textit{tr lt} & \textit{tr sn} & \textit{vegtr} & \textit{terr} & \textit{sky} & \textit{pers} & \textit{rider} & \textit{car} & \textit{truck} & \textit{bus} & \textit{train} & \textit{motor} &  \textit{bike} & \textbf{mIoU} \\
\hline
transfer & 96.15 &  72.94 & 86.60 & 30.17 & 37.85 & 33.60 & 38.27 & 54.37 & 88.24 & 50.20 & 90.41 & 64.34 & 30.64 & 87.71 & 27.18 & 41.42 & 25.13 & 22.37 & 61.49 & 54.68  \\
\hline
Self-learning & 97.60 &  81.12 & 90.53 &45.02 & 51.58 & $\mathbf{53.47}$ & 60.56 & 73.12 & 91.38  & 55.67 & 94.35 & 77.70 & 53.51 & 92.56 & 47.60 & 59.24 & 48.30 & 55.37 & 72.64 & 68.48  \\
\hline
STSO & $\mathbf{97.65}$ &  $\mathbf{81.77}$ & $\mathbf{90.70}$ & $\mathbf{51.52}$ & 51.64 & 53.45 & $\mathbf{60.75}$ & $\mathbf{73.48}$ & $\mathbf{91.46}$ & 58.46 & $\mathbf{94.46}$ & $\mathbf{78.08}$ & $\mathbf{55.18}$ & 92.55 & 45.30 & 59.68 & 43.45 & 53.17 & $\mathbf{72.92}$ & 68.72  \\
\hline
ATSO & 97.64 &  81.74 & 89.97 & 43.18 & $\mathbf{52.06}$ & 51.09 & 54.74 & 69.72 & 91.15 & $\mathbf{59.95}$ & 94.03 & 76.18 & 51.90 & $\mathbf{93.27}$ & $\mathbf{68.28}$ & $\mathbf{74.67}$ & $\mathbf{61.54}$ & $\mathbf{56.07}$ & 70.99 & $\mathbf{70.43}$  \\
\hline
\end{tabular} }
\caption{Class-wise and mean IOU (\%) of Cityscapes, produced by different training strategies.}
\label{tab:comparison_cityscapes}
\end{table}

\begin{figure*}[!b]
\vspace{-0.3cm}
\centering
\includegraphics[width=\textwidth]{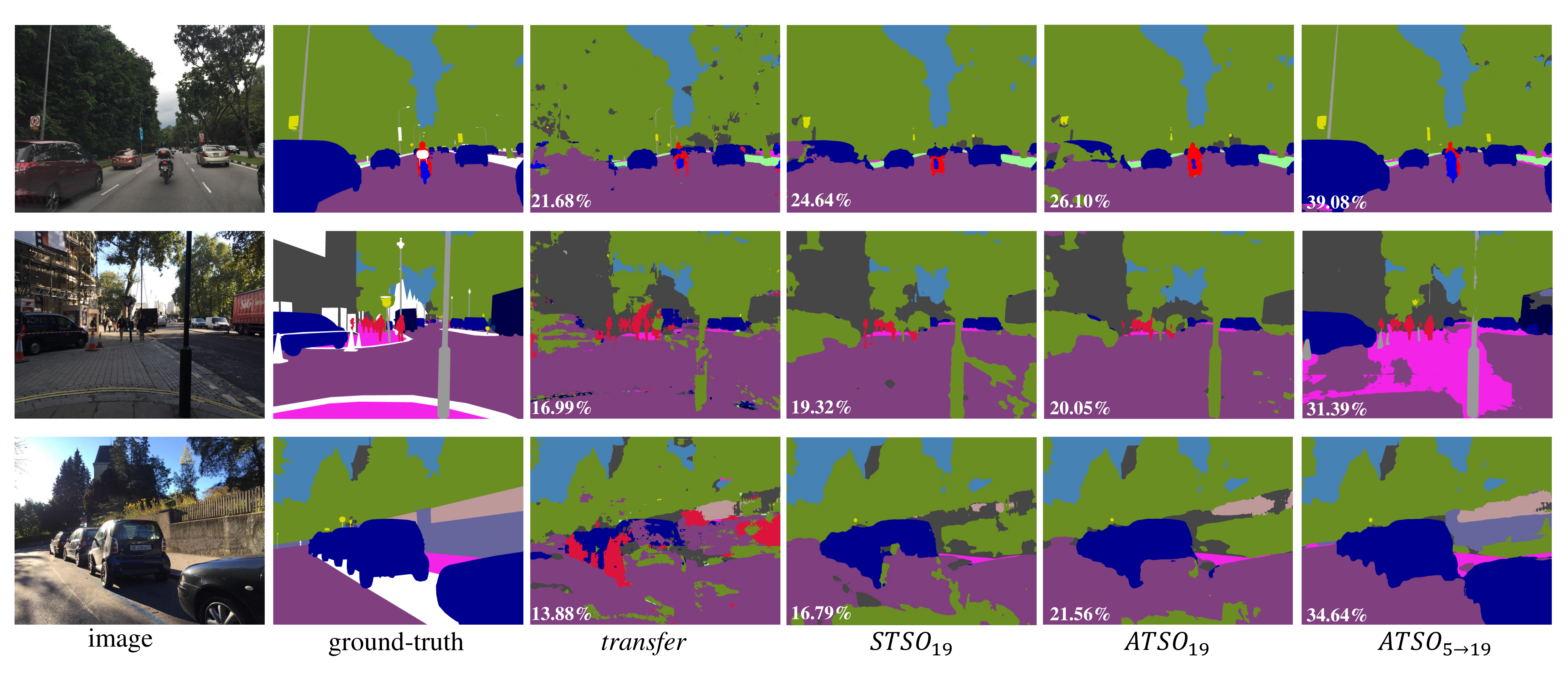}
\caption{Visualization of the results produced by the direct transfer baseline, STSO, ATSO, and tuning ATSO on the $19$-class pseudo label, respectively. Especially, some classes with a small area (\textit{e.g.}, \textit{traffic light}), are improved significantly in ATSO$_{5\rightarrow19}$. The number at the lower-left corner of each image indicates the IOU averaged over all classes -- although pixel-wise accuracy of ATSO seems good, this number can be impacted by some poorly segmented classes.}
\label{fig:visualization_mp2}
\end{figure*}

\section{Additional Results of Transfer Learning from Cityscapes to Mapillary}
\label{extra4}

This part corresponds to Section~\ref{transfer:natural} of the main article. First, we report the self-learning baseline for the transfer learning experiments from Cityscapes to Mapillary, which reports an accuracy of $26.97\%$ (averaged over all classes). This is done without using the pre-defined mapping to reduce the number of classes from $19$ to $5$. Note that this number is lower than the corresponding numbers of STSO ($28.11\%$) and ATSO ($28.26\%$).

To show how our algorithm improves segmentation, we provide some typical examples in the transfer learning task from Cityscapes to Mapillary. We find that the direct transfer results are often below satisfaction, but the semi-supervised learning approaches can improve domain transfer performance dramatically. In particular, ATSO works best among all the solutions. In the first and third example, we find that ATSO$_{5\rightarrow19}$ is more stable at producing good results for the large objects. This is mainly because the learning process is stabilized by the pseudo labels generated on the reduced $5$ classes. On the other hand, when we tune the model by generating the $19$-class pseudo label, the performance gain is often more significant on small objects. So, the proposed flowchart (starting from $5$ classes and then tuning on $19$ classes) is verified effective.

\end{document}